\def\year{2020}\relax
\documentclass[letterpaper]{article} 
\usepackage{aaai20}  
\usepackage{times}  
\usepackage{helvet} 
\usepackage{courier}  
\usepackage[hyphens]{url}  
\usepackage{graphicx} 
\usepackage{graphicx}  
\usepackage{amsmath,amssymb}
\usepackage{makecell}
\usepackage{dblfloatfix}
\usepackage[table]{xcolor}
\usepackage{soul}
\usepackage{graphics}

\usepackage[compact]{titlesec}

\title{Style Matters!\\Investigating Linguistic Style in Online Communities}
\author{Osama Khalid, Padmini Srinivasan\\ 
University of Iowa\\
\{osama-khalid, padmini-srinivasan\}@uiowa.edu 
}

\begin{document}

\maketitle

\begin{abstract}
Content has historically been the primary lens used to study language in online communities. This paper instead focuses on the linguistic style of communities. While we know that individuals have distinguishable styles, here we ask whether communities have distinguishable styles. Additionally, while prior work has relied on a narrow definition of style, we employ a broad definition involving 262 features to analyze the linguistic style of 9 online communities from 3 social media platforms discussing politics, television and travel. We find that communities indeed have distinct styles. Also, style is an excellent predictor of group membership (F-score 0.952 and Accuracy 96.09\%).
While on average 
it is statistically equivalent to predictions using content alone, it is more resilient to reductions in training data. 
\end{abstract}

\section{Introduction}
A key attraction in online social media is the capacity to establish specific communities (subreddits, subverses, etc.) for individuals with shared interests. An interesting aspect of 
these communities is that although their members may come from diverse locations and backgrounds, a shared system of language and communication allows them to engage effectively with each other. 
This shared communication system evolves naturally, i.e., \emph{in situ}, and is in turn also a part of what defines the community's identity -
an identity formed by highlighting the commonalities among group members and differences from other groups \cite{bucholtz2004language}. 
An example of this is the use of shibboleths as linguistic markers of identity \cite{ayuso2011lucky}.

A key component of a community's shared language is its vocabulary \cite{pennebaker2011secret} - undoubtedly influenced by shared interests. Members interact with each other using familiar content bearing words, phrases, symbols etc. In fact, a shared vocabulary is necessary to effectively engage within the community.

A second key component is the set of para-lingual or stylistic features. These can be explicit like the avoidance of taboo words or more subtle like the use of complex language. While individuals can consciously learn vocabulary to express content, style develops more through subconscious processes \cite{daelemans2013explanation}. 
It may be argued that language develops and is used in communities in a similar manner. 
A dynamic and subtle process of positive and negative feedback perhaps shapes a community's shared 
style over time.

\subsubsection{Style versus Content Research:}
Of the two, content has been studied extensively especially in the field of information retrieval. 

Individual writing style has been studied in domains like author attribution research \cite{safin2018detecting}.
We know that style can serve as a window into the psychological and sociological state of individuals and provide cues about their gender, occupation and even social class \cite{pennebaker2001linguistic}. In fact almost all of the research involving style, or stylometrics, has focused on the individual, leaving open a number of research questions about a community's style.
Another limitation in prior research is that researchers rarely maintain a clear distinction between content features and style features and often bundle them all under stylometry \cite{wang2017understanding,pennebaker2001linguistic}.
This seriously limits our understanding of the role of style in defining communities.

We see an important opportunity to study the para-language or stylistic features of communities.
It is known that groups can have a profound impact on individuals' identities \cite{abrams2006social}. The study of community-style can give us an additional lens to study the individual in the context of their relation to the community. 

\subsubsection{Research Goal:} Our goal is to study the linguistic style of 9 communities selected from voat, 4chan and reddit defined around the topics of politics, travel, and television.  We study them through a lens made of 262 style-features, taking care to avoid content revealing features.
Doing so, we ask the following novel research questions.

\noindent\textbf{RQ1:}
Do online communities have identifiable linguistic styles? (section ~\ref{subsection:community-style})

If the answer to the above question is yes, then our goal is to identify stylistic features that are distinctive both to a community and globally across communities. (section~\ref{subsection:distinctive-features})

\noindent\textbf{RQ2:} 
To what extent can we predict community membership based on style alone?  
Taking care to distinguish style from content, we compare prediction using style with a baseline strategy where prediction is done more traditionally using content 
(section~\ref{subsection:prection}).

In summary, we find that communities have representative and distinctive style features. 
Additionally, style predicts community membership with surprisingly excellent results ($>$95\% accuracy and $>$0.95 F-Score).
While this performance is on average statistically equivalent to content-based prediction, we observe that compared to content-based predictions, style-based predictions are more robust to a drastic reduction in training data.
In section~\ref{subsection:case-study} we present a case study comparing the styles of two communities.
Section 6 rounds out our study with additional analyses.  For example, we find that our style based classifier is excellent at predicting community membership even with thematically similar communities from the same social media platform.

\section{Related Research}

\subsubsection{Distinctiveness of style:}

In 2005, \citeauthor{van2005new} introduced the notion of a human `stylome' analogous to the human genome. 
This refers to distinctiveness in an individual's linguistic style similar
to an individual's distinct genetic makeup.
`Stylometrics' has been the basis of research on identifying author identity \cite{safin2018detecting,sapkota2014cross} 
and attributes such as gender, race and even social class (e.g., \cite{cheng2011author}). It has also been used to identify acquired characteristics such as political leanings \cite{potthast2017stylometric}. 

\subsubsection{Style as an indicator of psychological status:}

Pennebaker \shortcite{pennebaker2011secret} postulated that, because of the psycholinguistic nature of style, individuals who share similar mental illnesses would have a  similar language style. 
Using Flickr data, \citeauthor{wang2017understanding} \shortcite{wang2017understanding} employed style similarity to identify individuals susceptible to mental illnesses. \citeauthor{zaman2019detecting} \shortcite{zaman2019detecting} analyzed google search histories with LIWC features to find users with low self-esteem. 

\subsubsection{Studying style in groups:}

\citeauthor{hu2013dude}, \shortcite{hu2013dude} analysed the language style  of 3 broadly defined groups: Twitter, email and blogs.   
These groups, however, do not represent communities of individuals with shared interests, which is our focus.
\citeauthor{hu2016language}\shortcite{hu2016language} demonstrated that people with similar occupations share similar language style. 
However, individuals were selected based on their occupation and not on the basis of networking in `community' structures. 

The study that is closest in spirit to our work in being more community-focused is by \citeauthor {potthast2017stylometric}  \shortcite{potthast2017stylometric} analyzing the language of three groups of news articles (mainstream, and hyperpartisan right-wing and left-wing articles). 
They found similarities in the writing styles of both, extreme right-wing and left-wing articles; the mainstream articles had a different style. 
But here too, news articles are largely single direction transmitters of information with almost no support for interactivity as in communities.
Additionally, they use a limited set of style features.
\subsubsection{Limitations in prior work:}

(1) There has been no significant focus on examining style in communities; most of the work has focused on individuals and loosely formed groups (email/news articles).
(2)
With few exceptions such as \citeauthor{potthast2017stylometric}\shortcite{potthast2017stylometric} most papers include content features as part of style analysis.
This is not surprising since choice in vocabulary is recognized as a matter of style.
But, the downside is that such papers cannot clearly separate signals conveyed through content and 
those conveyed through style.
These limitations motivate our research.

\section{Methods}\label{methodology}

\subsection{Identifying a community's style}

A community's style may be understood via its representative stylistic features.

\subsubsection{Definition of Representative:}
A feature $f_{x}^c$ is regarded as more representative of a community $c$ than a feature $f_{y}^c$, if $f_{x}^c$'s values show greater consistency across $c$'s posts than $f_{y}^c$'s values.

More formally we assess the representativeness of feature $f_i$ for a community $c$ by the standard deviation ($\sigma$) of its values over $c$'s documents. 
Since features can have values with different ranges, these values are normalized  by scaling them between 0 and 1. 
\footnotesize
\begin{equation}
    \sigma_{f_{i}^c} = \sqrt{\frac{\sum_{j=1}^{M_c}f_{ij}^{c}-\overline{f_{i}^{c}}}{M_c-1}}
    \label{equation:eq1}
\end{equation}
\normalsize
Here $f_{ij}^{c}$ is the normalized feature value in pseudo-document $j$, $M_c$ is the number of pseudo-documents of community $c$, $\overline{f^{c}_{i}}$ is the mean of the scaled values of the feature. 
A pseudo-document is the concatenation of a collection of posts and is detailed in the next section.

\subsection{Identifying distinctive features}

There are two perspectives on identifying distinctive features.
First, a feature may be distinctive for a specific community and second, a feature may be distinctive across all communities.

\subsubsection{Definition of Distinctive for a Community:}
A feature $f_{x}^c$ is more distinctive from a community $c$'s perspective when compared to another feature $f_{y}^c$ if $c$'s average distance from all the other communities decreases by a larger extent upon exclusion of $f_{x}$ than upon exclusion of $f_{y}$.
\begin{table*}[h]
\centering
\scriptsize
\begin{tabular}{l|l|l|l|l|l}
\textbf{Social media} & \textbf{Community} & \textbf{Months} & \textbf{\# of comments} & \textbf{\# of comments} & Time Period \\ 
 & & &\textbf{ (Full set)} &\textbf{ (10k-subset set)} &   \\\hline

 4chan 1\% & /pol (/4c/politics) &63 & 1,697,788&10,265 & Nov'13 - Jan'19 \\
 & /trv (/4c/travel) & 59& 6,542& 6,542 &Feb'14 - Dec'18 \\
 & /tv (/4c/television) & 74& 777,126& 9,903 &Dec'12 - Jan'19 \\\hline
voat & /v/politics & 50& 862,501&10,073 & Jan'15 - Feb'19  \\
 & /v/travel &36 & 742&742  & Jul'15 - Feb'19 \\
 & /v/television & 50& 8,854&8,854 & Jan'15 - Feb'19  \\\hline
reddit 10\% & /r/politics & 77& 5,866,346&10,265 & Dec'12 - Apr'19  \\
 & /r/travel  &80& 312,862&10,037 &Dec'12 - Jul'19 \\
 & /r/television& 76& 830,068&9,931 &Jan'13 - Apr'19\\\hline
\end{tabular}
\caption{Summary of Datasets: Full set and 10k-subset.}

\label{table:summary}
\end{table*}

More formally we first define the distance ($d$) between two communities $i$ and $j$ as follows.
\footnotesize
\begin{equation}
    d(i,j) = \sqrt{\sum_{k=1}^N(\overline{f_{k}^i}-\overline{f_{k}^j})^2}\quad \quad  f\in F
    \label{equation:dist}
\end{equation}

\normalsize
where $F$ is the full set of features with dimension $N$.
We then define the average distance of a community $i$ from all other communities as:
\footnotesize
\begin{equation}
    \overline{d(i)} = \frac{\sum_{p=1}^{P}d(i,p)}{P}
    \label{equation:eq3}
\end{equation}
\normalsize
where $P$ is the number of communities.

Finally, distinctiveness ($\Delta C$) of a feature $\alpha$ for a community $i$ is defined as follows: 
\footnotesize
\begin{equation}
    \Delta C_i^\alpha=\overline{d(i)}-\overline{d^\alpha (i)}
    \label{equation:eq4}
\end{equation}
\normalsize
Where $\overline{d^\alpha(i)}$ is average distance (Eq.~\ref{equation:eq3}) computed on the feature space $F-\{\alpha\}$.

\subsubsection{Definition of Distinctive Globally:}
A feature $f_{x}$ is more distinctive from a global perspective when compared to another feature $f_{y}$  if the average distance between all pairs of communities decreases by a larger extent upon exclusion of $f_{x}$ than upon exclusion of $f_{y}$.

Distance between a pair of communities is as in Eq. \ref{equation:dist}.  The average is the mean distance across all pairs of communities.
We compute the average in two ways, once with all features $F$ and a second time with features in $F - \{\alpha\}$.  Their difference represents global distinctiveness and is labelled $\Delta G^\alpha$ for feature $\alpha$.

\subsection{Predicting community membership}

We approach this as a standard single label, multiple class (9 communities) classification problem. 
We represent each pseudo-document as a 262 dimensional vector of feature values. 
These features are described in the next section.

We use a random forest classifier\footnote{We also conducted experiments using a multi-class logistic regression model. However, the results were largely similar to results from the random forest classifiers, therefore we only report the results from the latter.} and analyze performance using accuracy, precision, recall and F-score.
We use 3 fold cross validation for our prediction experiments ensuring that all comments from the same user (where the user-id is persistent) fall into the same fold.

\subsubsection{Baseline:}
We use content-based classification as a baseline predictor. We represent each pseudo-document (described later) as a weighted vector of words after stemming and excluding stop words and purely numeric unigrams. The stems are given TF-IDF weights. 
We use identical 3 fold cross validation and random forest classifier, to calculate the accuracy, precision, recall and F-score.

\section{Dataset and Features}\label{sec:data}
\subsection{Dataset}

\begin{table*}[]
\centering
\scriptsize
\begin{tabular}{l|l|l|l}
\textbf{ Readability} & \textbf{ \makecell[l]{Word level\\ features}} & \textbf{  \makecell[l]{Parts of Speech \\ Frequencies}} & \textbf{ \makecell[l]{Character \\Frequencies}} \\\hline

Automated Readability Index (ARI)\scriptsize{(D}) & Hapax (Dis)Legomena (D) & Verbs\scriptsize{(PWS)} & Emoji \\
Coleman-Liau Index (CLI)\scriptsize{(D)} & Brun\'et's W Measure\scriptsize{(D)} & Conjunctions\scriptsize{(PWS)} & White Space\scriptsize{(PWSC)} \\
Dale Chall Readability Index\scriptsize{(D)} & Yule's K Characteristic\scriptsize{(D)} & Determiners\scriptsize{(PWS)} & Digits\scriptsize{(PWSC)} \\
Simple Measure of Gobbledygook (SMOG)\scriptsize{(D)} & Honore's R Measure\scriptsize{(D)} & Existentials\scriptsize{(PWS)} & Tabs\scriptsize{(PWSC)} \\
Flesch Kincaid Grade Level\scriptsize{(D)} & Sichel's S Measure\scriptsize{(D)} & Prepositions\scriptsize{(PWS)} & Special Characters\scriptsize{(PWSC)} \\
Fleash Kincaid Readability Ease\scriptsize{(D)} & Simpson's Index\scriptsize{(D)} & Adjectives\scriptsize{(PWS)} & Uppercase\scriptsize{(PWSC)} \\
Gunning Fog Index\scriptsize{(D)}& \makecell[l]{Out of Vocabulary Rate\scriptsize{(D)}} & Nouns\scriptsize{(PWS)} & Linebreaks\scriptsize{(PWSC)} \\
 & Syllable Frequency\scriptsize{(PWS)} & Possessives\scriptsize{(PWS)} & Punctuations\scriptsize{(PWSC)} \\
 & Short Word Frequency\scriptsize{(PWS)} & Adverbs\scriptsize{(PWS)} & Character Count\scriptsize{(PWS)} \\
 & Elongated Words\scriptsize{(PWS)} & Possessives\scriptsize{(PWS)} & \\
 & LIWC-$v3$\scriptsize{(PWS)} & Interjections\scriptsize{(PWS)} & \\
 \hline
\end{tabular}

\caption{Categories of Style Features. Granularities: D=Document, P=Post, W=Word, S=Sentence, C=Character}

\label{tab:featsAll}
\end{table*}
We collected comments posted in 9 communities discussing politics, travel and television from 4chan\footnote{www.4chan.org}, reddit\footnote{www.reddit.com} and voat\footnote{www.voat.vo}.

\begin{table*}[b!]
\centering
\scriptsize
\begin{tabular}{c|c|c|c|c|c}
\multicolumn{2}{c|}{\textbf{LIWC-$v1$}} & \multicolumn{2}{c|}{\textbf{LIWC-$v2$}} & \multicolumn{2}{c}{\textbf{LIWC-$v3$}}  \\\hline
\textbf{Categories}&\textbf{Subcategories}&\textbf{Categories}&\textbf{Subcategories}&\textbf{Categories}&\textbf{Subcategories}\\\hline
&\scriptsize Function Words& &\scriptsize Function Words& &\scriptsize Function Words\\
\scriptsize Linguistic Processes&\scriptsize Common Verbs&\scriptsize Linguistic Processes&\scriptsize Common Verbs&\scriptsize Linguistic Processes&\scriptsize Common Verbs\\
&\scriptsize Swear Words& &\scriptsize Swear Words& &\scriptsize Swear Words\\\hline
&\scriptsize Social Processes& &\scriptsize Social Processes & &\scriptsize Cognitive Processes\\
&\scriptsize Affective Processes & &\scriptsize Affective Processes & &\\
\scriptsize Psychological Processes& \scriptsize Cognitive Processes &\scriptsize Psychological Processes &\scriptsize Cognitive Processes &\scriptsize Psychological Processes &\\
& \scriptsize Perceptual Processes & &\scriptsize Perceptual Processes & &\\
& \scriptsize Biological Processes & & & &\\
& \scriptsize Relativity & & & &\\\hline
\scriptsize Spoken Concerns & & \scriptsize Spoken Concerns & & \scriptsize Spoken Concerns &\\\hline
\scriptsize Personal Concerns & & & & &\\\hline

\end{tabular}
\caption{The categories and subcategories of features in each version of LIWC}
\label{tab:LIWC}

\end{table*}

\subsubsection{reddit:}
Reddit is a news aggregation website with over 1 millions subreddits; specific communities with shared interests. 
Users interact with others by commenting on posts. 
The Reddit API for data collection has multiple limitations including being rate-limited\footnote{https://github.com/reddit-archive/reddit/wiki/api}. 
Instead, we obtained our data from pushshift.io\footnote{https://github.com/pushshift/api}, a reddit archive.  
We downloaded comments from /r/politics, /r/travel and /r/television subreddits going back to Dec. 2012. 
We further reduced the large data volume by taking a 10\% sample of the data for each community.

\subsubsection{voat:}

Voat is an alt-right variant of reddit \cite{reynolds_2018}.
Similar to subreddits, voat has ``subverses''.
However, unlike reddit, voat has neither a well-developed API nor an archive. We manually scraped the comments 
from the three voat communities /v/politics, /v/travel and /v/television. Our data spanned from Jan. 2015 to Feb. 2019.

\subsubsection{4chan:}
4chan is an image-board  website \cite{bernstein20114chan}, where users post anonymously. Like subreddits and subverses, most of its 63 boards are thematic in nature.
Unlike reddit or voat, 4chan is completely anonymous; pseudonyms are optional but rare.
We assume that each comment is posted by a unique user.
We downloaded the political discussions (/pol), the travel (/trv) and television(/tv) boards data from an archiving service, 4plebs\footnote{http://archive.4plebs.org/} which goes back to Dec. 2012. 
We label these as /4c/politics. /4c/travel and /4c/television. 
Like reddit, 4chan also has an extremely high volume of data.
We use a 1\% sample of each board. Table \ref{table:summary} summarizes our full dataset and a reduced 10k-subset. Unless otherwise specified all results are on the full dataset.

\subsubsection{Pseudo-documents:}
Our goal is to study the style of a community and not of its specific individuals. 
Thus, we disregard user identity and create pseudo-documents 
containing temporal chunks of comments. 
These pseudo-documents are representative of the community and used to train and test our community-level style and content classifiers. 
Each pseudo-document holds all comments posted in a single month by a community.
As an example, for /4c/politics there are 63 pseudo-documents corresponding to 63 months of data.
For some communities, the data can be sparse for a given month, as an example, only 4 comments were posted on the /v/travel forum during Jan. 2016.

\subsection{Features}
Most prior stylometry projects, such as \citeauthor{feng2012syntactic} \shortcite{feng2012syntactic}, have relied on narrow definitions of style. Since there is no \emph{a priori} theoretical reason for selectivity in style features we use a wide array of features identified from prior literature. 
Crucially, we exclude features that might also convey information about topic  or content. Table \ref{tab:featsAll} presents our four style categories.

\subsubsection{Readability:} These measure the ease with which one can expect a text to be comprehended. Included are features such as CLI \cite{coleman1975computer}, and the Gunning fog index \cite{gunning1969fog}. 
Prior literature, including \citeauthor{potthast2017stylometric} \shortcite{potthast2017stylometric}, have used a similar set of readability features for stylometric analysis.

\subsubsection{Parts of Speech:} These capture syntactic properties of style by calculating the distributions of various parts of speech like nouns and verbs. 

\subsubsection{Character level features:} These measure  orthographical style properties and include features such as the use of white space, punctuation, and emojis.

\subsubsection{Word level features:} These assess the diversity and range in vocabulary used (but the specific words appearing in the text are not included as features). 
We include LIWC in this category as its logic is mostly word dictionary driven, though it also includes measures for properties such as short word counts. 
LIWC measures the proportion of words which fall into  one or more hierarchical categories and their subcategories \cite{tausczik2010psychological}.

Some LIWC categories have the capacity to \emph{leak} information about the text's topical content.
We take precautions to avoid including such features when building style-based classifiers. We do this in order to make a fair comparison between style and content for predicting community membership.
For example, if a document scores high on the LIWC category `Religion' this can indicate that the posts are related to religion.
Likewise, the category `Personal Concerns' has subcategories such as `money' which includes words related to finance. Again, these potentially indicate post topic.

We create two reduced versions of LIWC which exclude topic leaking categories.
In LIWC-$v2$, we remove 9 features belonging to `Psychological Processes' category (e.g., `ingestion' which includes words like eat, pizza etc) and 7 from the `Personal Concern' category (e.g., `religion') as
these almost certainly convey topic.
In LIWC-$v3$, we further eliminate 10 features which we suspect \emph{indirectly} leak  content -  such as sentiment 
(under `Affective Processes' category) and `Perceptual Process' categories (words like touching).
Again we take these precautions in order to get a clearer understanding of style versus content in communities.
We suggest that this level of caution in avoiding inclusion of topical features in style is one of the strengths of our work.   
The full LIWC (LIWC-$v1$) has 64 features while LIWC-$v2$ and our most conservative LIWC-$v3$ have 48 and 38 features respectively.
Table \ref{tab:LIWC} summarizes the features retained in each LIWC version.
We use our most conservative LIWC-$v3$ unless otherwise specified. 

We measure style features at multiple granularities: word, sentence, post and (pseudo-) document levels as appropriate.
For example, the prevalence of conjunctions may be the average of frequencies across posts or sentences or a single proportion of the total words in the document.
Some measures are only suitable at the document level.  
Our final feature set, with LIWC-$v3$, has 262 features in total. Table \ref{tab:featsAll} summarizes the features we explore and their granularity.

\section{Results}\label{results}

\begin{table*}[tp!]
\centering
\scriptsize
\resizebox{2.1\columnwidth}{!}{
\begin{tabular}{r|p{1.16cm}p{1.16cm}p{1.16cm}|p{1.16cm}p{1.16cm}p{1.17cm}|p{1.17cm}p{1.17cm}p{1.17cm}}
&& \textbf{ politics} &  &  & \textbf{ travel} &  &  & \textbf{ television} &  \\\hline
&\textbf{voat} & \textbf{4chan} & \textbf{reddit} & \textbf{voat} & \textbf{4chan} & \textbf{reddit} & \textbf{voat} & \textbf{4chan} & \textbf{reddit} \\\hline
&SYM & swear & RB & simpson & yule & pronoun & auxverb & negate & totChar \\
&you & inhib & VB & tabs & simpson & IN & FW & adverb & CLI \\
&nonfl & MD & DT & RBS & VB & TO & VBG & excl & charWord \\
&digits & we & ipron & filler & upper & adverb & NN & preps & WRB \\
&WDT & EX & funct & SYM & NNP & VBG & emoji & IN & alphabet \\
&UH & JJS & VBN & WP\$ & VBG & preps & upper & VBG & ARI \\
&assent & ppron & punctuations & emoji & swear & past & VBP & short & MD \\
&VBN & shehe & VBZ & assent & you & VBN & verb & WRB & VBN \\
&auxverb & you & PRP & shehe & VBP & NNS & NNS & PRP\$ & CC \\
\textbf{Range}&cause & assent & NN & JJR & lines & VBD & VBN & shehe & VBP\\\hline
Top 10&{\scriptsize(0.085-0.105)}&{\scriptsize(0.058-0.072)}&{\scriptsize(0.052-0.059)}&{\scriptsize(0.096-0.100)}&\scriptsize{(0.067-0.075)}&{\scriptsize(0.056-0.062)}&{\scriptsize(0.081-0.094)}&{\scriptsize(0.053-0.060)}&{\scriptsize(0.055-0.058)}\\\hline
Full set&{\scriptsize(0.085-0.224)}&{\scriptsize(0.058-0.241)}&{\scriptsize(0.052-0.208)}&{\scriptsize(0.096-0.312)}&{\scriptsize(0.067-0.281)}&{\scriptsize(0.056-0.210)}&{\scriptsize(0.081-0.239)}&{\scriptsize(0.053-0.246)}&{\scriptsize(0.055-0.180)}\\
\hline

\end{tabular}
}
\caption{Top ten representative features for each community.}
\label{table:repFeat}
\end{table*}
\subsection{The Style of a Community}
\label{subsection:community-style}

The style of a community is represented by its dominant stylistic features.
Table \ref{table:repFeat} presents the features with the top ten representative scores (Eq.~\ref{equation:eq1}) for each community. 
Since features calculated at multiple granularities in spirit aim at the same stylistic property, we rank feature types based on the best representativeness score across granularities. The table shows the range of standard deviations as well.

Most (42 of 61, 69\%) of the top 10 representative features apply to a single community.
12 features (20\%) apply to 2 communities.
Verb feature tend to occur commonly, such as `VBN' (past participle verbs) appearing in five communities and `VBG' (participle verbs) appearing in 4 communities.

\subsubsection{Standard deviations are low:} 
In general, the standard deviations of the ten most representative features, for all communities, have fairly narrow ranges staying mostly between a low of 0.053 to a high of less than 0.1. 
Thus these top properties are fairly consistent across the pseudo-documents of a community.
Voat communities had the highest range while Reddit communities tended towards the lowest ranges.
As expected the standard deviation ranges increase when considering all features as shown in the table.

\subsubsection{Representative features within each category:}
Here we identify the single feature, within each of the four categories, that has the best rank for representatives on average across the 9 social media.
Table~\ref{tab:repCatRank} provides the  mean score and standard deviation for these top features. We also include information for LIWC-$v3$.

An interesting feature here is `CLI' under readability. A score of 6.98 would indicate that the language of social media is simple enough that an individual with almost 7 years of education can understand it. The `CLI' was the lowest for /4c/television at 5.31 and the highest for /r/politics at 8.81. This would indicate that a Middle School freshman can understand the language on /4c/television, while an individual would need to be almost a high school freshman to understand the language on /r/politics.

\begin{table}
\centering
\scriptsize

\begin{tabular}{l|l|l}
\textbf{Category}& \textbf{Feature}&\textbf{Mean($\sigma$)}\\ \hline \hline
\textbf{Readability} & CLI&6.98(0.967)\\
\textbf{Words} &simpson&0.992(0.004)\\
\textbf{POS} & VBG/Word&0.025(0.002)\\
\textbf{Character}&word length&4.26(0.115)\\ \hline
\textbf{LIWC-$v3$} & ppron/words&0.0421(0.006)\\\hline

\end{tabular}
\caption{Representative feature in each category.}
\label{tab:repCatRank}

\end{table}
\subsubsection{Styles are different across communities.}
\begin{figure}[!b]
    \centering
    \includegraphics[width=0.95\columnwidth]{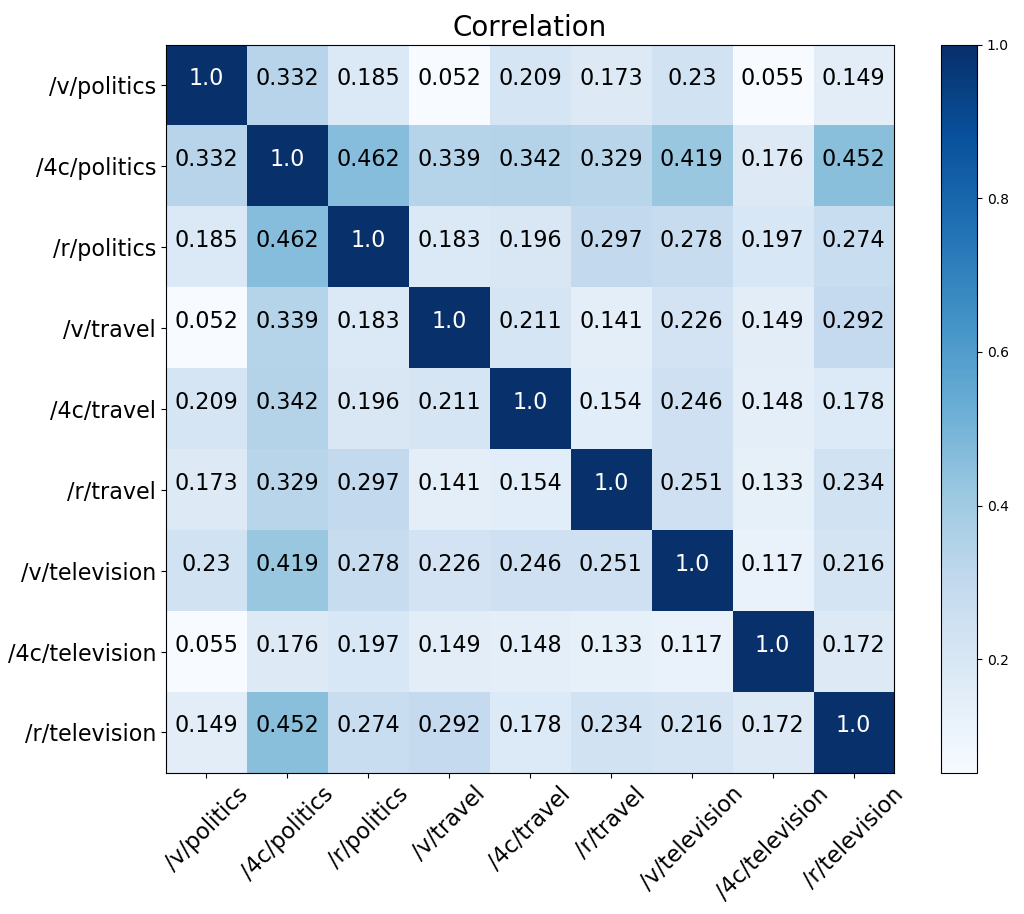} 
    
        \caption{Spearman correlation between communities. Each community is represented by a ranking of features by representativeness score.}
            \label{fig:corr}
\end{figure}

We rank all the style features by the representative score for each community 
and compare communities to identify style similarities using Spearman's rank-order correlation (Fig. \ref{fig:corr}).
17 of 36 community pairs have weak correlation (less than 0.2) while 15 pairs are weak to moderate correlations (greater than 0.2 and less than 0.4).
The three highest correlations, while still at best moderate, all involve /4c/politics. This would indicate that these communities share representative style features - but only to a slight extent.

\subsubsection{Takeaway: }
Our answer to \textbf{RQ1} is that communities do have their own style which is identifiable and that style differs across communities independent of topic and medium.

\subsection{Distinctive Style Features}
\label{subsection:distinctive-features}

\subsubsection{Community specific distinctive features:}

Table \ref{table:disFeat}, shows the top ten features that make each community distinct.
Again we consider the best distinctiveness score for a feature type independent of granularity.  
The table also shows the range of distinctiveness score for the feature ($\Delta C$ score).
This ranges from 0.075 to 0.178.

Of the 49 distinct features in the top 10 lists, 23 (47\%) and 17 (35\%) were listed for one or two communities respectively. 
The use of the present tense was the most common distinctive feature in 5 communities.
Brun\'et's W Measure was one of two appearing as distinctive for 4 communities.

\subsubsection{Globally distinctive features:}

Table \ref{table:disFeat}, also shows the most and least globally distinctive features ($\Delta G$ score). 
Verb categories, Brun\'et's W Measure and insight are amongst the most distinctive globally.
As expected when we rank all features by the standard deviation of their mean scores across communities, 9 of the 10 highest deviation features are amongst the most globally distinct.
These deviations range from 0.2445 to 0.2673. 
The one exception, auxiliary verbs (auxverb) - the 9th most distinct feature - was  ranked 12 by deviation.

\subsubsection{Consistency between community-specific and globally distinct features.}

\begin{table*}
\begin{center}

\scriptsize
\begin{tabular}{l|p{0.6cm}|p{0.6cm}|p{0.6cm}|p{1.1cm}|p{1.35cm}|p{0.6cm}|p{0.7cm}|p{0.6cm}|p{0.6cm}||p{0.7cm}|p{0.8cm}}
& \multicolumn{9}{c||}{\textbf{Community Specific, $\Delta C$}} &  \multicolumn{2}{c}{\textbf{Global, $\Delta G$}} \\ \hline
&\multicolumn{3}{c}{\textbf{politics}}  & \multicolumn{3}{|c|}{\textbf{travel}}  & \multicolumn{3}{c||}{\textbf{television}} & \multicolumn{2}{c}{} \\\hline

&\textbf{voat}& \textbf{4chan}& \textbf{reddit} &\textbf{voat}& \textbf{4chan}& \textbf{reddit}&\textbf{voat}& \textbf{4chan}& \textbf{reddit}&\textbf{Most}&\textbf{Least}\\\hline

&VBP& they& VBP & oov5000& verb& TO  &oov5000& CD& RBR&VBP&lines \\
&brunet & WP & honore &oov1000& present& assent &auxverb& CC& insight&verb&tabs \\
&you & UH& insight& oov500& honore& preps &VBP& totChar& TO&brunet&emoji \\
&swear& brunet& VBZ &hapax Leg.& article& VB &verb& EX& we&oov5000&LS\\
&oov500& verb& they &brunet& VBP& CD  &future& present& certain&oov500&FW\\
&CLI& VBZ& JJ&NNS& adverb& oov500&JJ& auxverb& WP&WP&yule\\
&shehe& auxverb& preps&future& CLI& inhib&article& verb& adverb&oov1000&simpson\\
&EX& present& ppron&MD& NN& present  &brunet& shehe& upper&present&elongation \\
&ppron& oov500& WP\$&quant& hapax Disleg.& MD  &i& VBZ& article&auxverb&filler \\
\textbf{Range}&present& NNP& VBG &WP& they& EX &past & SMOG & NNP&insight&NNPS \\\hline
Top 10&\scriptsize{(0.118-0.105)}&\scriptsize{(0.092-0.075)}&\scriptsize{(0.116-0.097)}&\scriptsize{(0.148-0.132)}&\scriptsize{(0.090-0.081)}&\scriptsize{(0.103-0.092)}&\scriptsize{(0.134-0.119)}&\scriptsize{(0.178-0.166)}&\scriptsize{(0.165-0.156)}&\scriptsize{(0.019-0.016)}&\scriptsize{(0.0001-0.0033)}\\ \hline

\end{tabular}

\caption{Top ten distinctive features and scores for each community.}
\label{table:disFeat}
\end{center}
\end{table*}

\begin{table*}[b!]
\centering
\scriptsize
\begin{tabular}{l|cccc|cccc}
\textbf{Community} & \multicolumn{4}{c}{\textbf{Style}} &  \multicolumn{4}{|c}{\textbf{Content}}  \\\hline
&Accuracy(\%)&Precision&Recall&F-score&Accuracy(\%)&Precision&Recall&F-score\\\hline
voat&92.10&0.901&0.921&0.911&85.47&0.832&0.855&0.843\\
\qquad/v/politics&98.00&0.987&0.980&0.983&92.00&0.993&0.920&0.955\\
\qquad/v/travel&83.96&0.802&0.840&0.820&77.57&0.654&0.776&0.709\\
\qquad/v/television&91.95&0.890&0.919&0.904&84.56&0.834&0.850&0.840\\\hline
4chan 1\%&97.93&0.983&0.979&0.981&97.27&0.953&0.973&0.963\\
\qquad /4c/politics&98.95&0.979&0.989&0.984&96.83&0.989&0.968&0.979\\
\qquad /4c/travel&97.72&0.972&0.977&0.975&97.73&0.869&0.977&0.920\\
\qquad /4c/television&97.22&0.995&0.972&0.984&97.30&1.00&0.973&0.986\\\hline
reddit 10\%&96.87&0.979&0.969&0.974&94.99&0.982&0.950&0.966\\
\qquad/r/politics&95.19&0.985&0.952&0.968&96.10&0.987&0.961&0.974\\
\qquad/r/travel&97.87&0.966&0.979&0.973&93.33&0.966&0.933&0.949\\
\qquad/r/television&97.37&0.987&0.974&0.980&95.61&0.995&0.956&0.975\\ \hline
Average&96.08&0.951&0.954&0.952&93.50&0.921&0.923&0.922\\
\hline
\end{tabular}
\caption{Performance on predicting community membership (Full set).}
\label{table:groupAcc}
\end{table*}

Not surprisingly, there are consistencies. For example, the most frequently occurring community-level distinctive feature is the use of present tense which is also globally distinctive. 
Similarly, globally the most distinctive feature, the use of present tense singular verbs (VBP) is also distinctive for 4 communities.
At the tail end, the use of linebreaks is least distinctive (see Table \ref{table:disFeat}).
It is also least distinctive for 8 of our 9 communities (not shown in tables). 

\subsubsection{Difference between distinctive and representative features.}  
We observe that features which are extremely representative of communities generally are not distinctive, either globally or at the community level.
Features such as Yule's K Measure were in the top 10 representative features for /4c/travel and /v/travel, however, globally they were amongst the least distinctive features. Even at a community level, Yule's K Measure is not a distinctive feature, ranking as the 12th and 18th \emph{least} distinct feature for /4c/travel and /v/travel respectively.
Similarly, while the use of swear words is a representative feature for /4c/politics and /4c/travel (Table \ref{table:repFeat}), it is not a distinct feature for either community. Instead, it is a distinct feature for /v/politics. 
These indicate that representativeness and distinctiveness are different properties.

\subsubsection{Takeaway:}
In summary, we find that there is consistency between community-specific and globally distinct features. And representativeness and distinctiveness appear to capture different properties.

\subsection{Case Study:}
\label{subsection:case-study}

The two thematically similar communities /r/television and /4c/television have vastly different discourse styles. Here we analyse their language complexity and their use of negations. 

/4c/television has the lowest average `CLI' of all communities at 5.31, whereas /r/television has a relatively high average `CLI' of 7.44. Compared to a user on /4c/television, a user on /r/television would require an additional 2 years of education to understand the discourse. Similarly, the `SMOG' score (19.4 versus 14.5) and `ARI' (5.4 versus 2.4) also indicate that the language on /r/television is more complex than on /4c/television.

Negations, on average, occur once every 4 sentences on /r/television, and less frequently, once every 8 sentences in /4c/television. The standard deviation of negations, for /4c/television is half that for /r/television, indicating more consistency in the rarer use of negations for the 4chan community. 
Drawing from Gonzales et al. \shortcite{gonzales2010language} which indicates that a higher use of negations within a community is indicative of lower cohesion between its members, we suggest that there is less disagreement amongst /4c/television users.
This in turn might indicate a stronger shared group identity amongst 4chan users.

This notion is further supported by differences in the use of first-person pronouns. 
We expect communities with a stronger sense of group identity to use these less frequently than communities where there is more individualism.
We observed that users are 1.7 times more likely to use the first person pronoun on /r/television than on /4c/television.
Possibly the increased emphasis on individualism also leads to greater disagreement (negation) on /r/television compared to /4c/television.

In sum, we note that users on /4c/television appear to have a stronger sense of group identity and they express themselves using a  less complex language.
This case study illustrates that by analyzing style we can gain deeper insights about specific communities and the differences between them.

\subsection{Predicting Community Membership}
\label{subsection:prection}

Table \ref{table:groupAcc}, presents the results of our prediction experiments addressing our research question \textbf{RQ2}.

\subsubsection{Style is an excellent predictor of community membership.}

With the exception of /v/travel, we observe accuracies higher than 90\% and F-scores above 0.9.
In almost all cases style results are numerically about the same or better than content for both accuracy and F-score.
The biggest wins are seen in voat. 
For example, content (0.71 F-score) has an even harder time predicting for /v/travel than style (0.82 F-score).
Community-level average scores are 0.95 for 4chan and reddit and above 0.90 for voat. 
While we expected some degree of success these strong results exceed our expectations.
These again indicate that style is distinct to each community.
 Comparing average scores (last row of Table \ref{table:groupAcc}) we find content and style to be statistically equivalent (p>0.05)\footnote{We use Smucker et al.'s \shortcite{smucker2007comparison} bootstrap test of significance in this paper.}.
Thus style and content are equally excellent at predicting community membership.
Thus, our answer to \textbf{RQ2} is that style excels in predicting community membership.

We note that substituting LIWC-$v2$ for LIWC-$v3$ drops performance an insignificant amount (accuracy 95.84, F-score 0.95). 
This tells us that prediction performance is unperturbed if we ignore the `Affective' and `Perceptual' process categories of LIWC.  
In particular this includes sentiment, commonly used in prior style research.

\begin{table*}[]
\centering
\scriptsize
\begin{tabular}{l|ll|ll|ll|ll}
\textbf{Community}  &  \multicolumn{2}{c}{\textbf{Style}} & \multicolumn{2}{|c}{\textbf{Drop}} & \multicolumn{2}{|c}{\textbf{Content}} & \multicolumn{2}{|c}{\textbf{Drop}} \\\hline
&Acc.(\%)&F1&\makecell[l]{\%$\Delta$ Acc.}&\makecell[l]{\%$\Delta$ F1}&Acc.(\%)&F1&\makecell[l]{\%$\Delta$ Acc.}&\makecell[l]{\%$\Delta$ F1}\\\hline

voat&78.57&0.817& -14.69 &-10.32 &87.71&0.878&2.63 &4.15 \\
\qquad/v/politics&90.48&0.884& -7.68 &-10.12 &90.48&0.896&-1.66 &-6.22 \\
\qquad/v/travel&62.73&0.704&-25.29 & -14.17&88.29&0.831& 13.82 & 17.07 \\
\qquad/v/television&78.52&0.823&-14.60 &-8.89 &84.56&0.900&0 &7.14 \\\hline

4chan 1\%&88.60&0.886& -9.53 & -9.53&69.23&0.742&-28.8 &-22.95 \\
\qquad/4c/politics&89.73&0.855&-9.31 & -13.06&69.19&0.703&-28.54 & -28.13\\
\qquad/4c/travel&89.77&0.916&-8.14 &-6.01 &68.18&0.734&-30.23 &-20.20 \\
\qquad/4c/television&86.70&0.894& -10.83 &-9.15 &70.18&0.785&-27.87 &-20.45 \\\hline

reddit 10\%&92.89&0.907& -4.11 & -6.89&91.57&0.866&-3.61 &-10.36 \\
\qquad/r/politics&94.14&0.915& -1.10 & -5.53&95.05&0.885&-1.10 & -9.14\\
\qquad/r/travel&93.01&0.924&-4.96 &-4.98 &86.90&0.834&-6.89 &-12.09 \\
\qquad/r/television&91.52&0.882&-6.01 &-10.04 &92.89&0.878&-2.85 &-9.97 \\\hline
Average &87.89&0.868& -7.3 & -8.83 &82.85&0.831& -9.09 &-9.97 \\ \hline
\end{tabular}
\caption{Performance on predicting community membership (10k subset).}

\label{table:group10kAcc}
\end{table*}

\subsubsection{Analysis of misclassifications:}

The biggest losses were from a content perspective: /v/travel was misclassified as /v/television (8.41\%) as /r/travel (7.48\%) and as /4c/travel (5.61\%).
The topical overlap between /v/television and /v/travel and between the various travel communities seems to be challenging for content-based prediction.
Style also confused /v/travel but the errors were fewer. 
Style mapped /v/travel to /v/television(7.56\%), /r/travel(3.77\%) and /4c/travel(0.94\%).
The style confusion between /v/travel and /v/television could be because of style similarities at the medium level, an aspect to probe in further research.
As an aside, the politics communities are highly distinguishable across media via content which leads us to infer that they are sufficiently different in topic.

\subsubsection{Results on a 10k-subset:}

A possible explanation for the relatively weak results for content-based prediction in /v/travel could be the low volume of data (only 742 comments, Table \ref{table:summary}).
All other communities had at the least 5,000 comments or greater, with a majority having more than 100,000 comments.
To test this, we repeated the prediction experiments by downsampling each dataset to around 10K posts (10k-subset).
If our intuition is supported then content based performance should improve for /v/travel.
With style classifiers, it is unknown what is likely to happen.

Table \ref{table:summary} describes the 10k-subset
and Table \ref{table:group10kAcc} compares performance between the full and the 10k-subset experiments.
When sizes are more comparable, we observe 14\% and 17\% improvements in accuracy and F-score respectively for content-based classification in /v/travel. 
Likewise, content performance also improves (but for F-score only) for /v/television - the community with the second smallest dataset.
 These support our intuition that content based prediction for voat was limited in the original experiment by the small dataset size (relative to the other communities).

Interestingly, the content classifier degrades markedly for 4chan in the reduced dataset, but not as much for reddit.
Taking the politics communities as an example, we note that in the full set, 
/4c/politics shared only 1.90\% of its vocabulary with the /4c/television.
In contrast, /r/politics shared 11.62\% of its vocabulary with /r/television.
In the 10k set, the former percentage increased by almost 11 times compared to just an increase of 2 times for the latter.
This observation is consistent with the classic guideline that training data size matters when building content based classifiers.

In contrast, for style-based prediction, this dependence is not as severe. 
While performance drops with style for all communities, 4chan style based prediction is markedly more resilient to downsizing than 4chan content based prediction.
With reddit, the drops in style-scores and content-scores are about even across the two measures.
Overall, style performance drops are less for style (7.3\% in accuracy and 8.83\% in F-score) than for content (9.09\% accuracy and 9.91\% F-score).
However, statistical tests on the average scores indicate once again that style and content are equivalent in predicting community membership.

\subsubsection{Takeaway:}

The answer to \textbf{RQ2} is that prediction based on style gives excellent results which are on average statistically equivalent to  
prediction using content.

Additionally, style classifiers appear less sensitive to reductions in training data.

\section{Additional Analysis \& Discussion}

\begin{table*}[t]
\centering
\scriptsize
\begin{tabular}{l|cccc|cccc}
\textbf{Window Size} & \multicolumn{4}{c}{\textbf{Style}} &  \multicolumn{4}{|c}{\textbf{Content}}  \\\hline
&Accuracy(\%)&Precision&Recall&F-score&Accuracy(\%)&Precision&Recall&F-score\\\hline

Half Month&93.81&0.916&0.908&0.912&93.08&0.911&0.914&0.912\\
\textit{One Month}&\textit{96.08}&\textit{0.951}&\textit{0.954}&\textit{0.952}&\textit{93.50}&\textit{0.921}&\textit{0.923}&\textit{0.922}\\
Two Month&94.56&0.937&0.930&0.933&93.14&0.923&0.926&0.925\\
\hline
\end{tabular}
\caption{Performance on predicting community membership for varying time windows.}

\label{table:windowAcc}
\end{table*}
\subsection{Feature Ablation Analysis}

We find that even when ignoring any one of the four style categories listed in Table \ref{tab:featsAll}, style on average numerically outperforms the content-based model (with $>$1.2 million features).  
The lowest performance obtained was when we excluded all LIWC-$v3$ features and used the remaining set of 148 features (accuracy 94.99\%, F-score 94.18). Table \ref{table:exclusive} summarizes these results.

When used individually the word-level features (135 features) and LIWC-$v3$ (114 features) are the only ones to numerically beat content. 
The worst performer is readability (6 features), but the scores are still good:  accuracy 86.97\%, F-score  84.40. Table \ref{table:inclusive} summarizes these results with full feature set results given in the first two rows.

\subsection{Effect of Window size on Results}
In our main results we had used 1-month as the window length to create pseudo-documents. 
Table \ref{table:windowAcc}
shows the effect of using 2-month and half-month windows compared to the one-month windows.  
It can be seen that compared to the content classifier, the accuracy of the style classifier exhibits slight variations. However, even with varying window sizes, the style and content classifiers continue to be statistically equivalent.

\begin{table}[h]
\centering
\scriptsize
\begin{tabular}{p{0.85cm}|p{1.1cm}|p{0.7cm}|p{0.9cm}|l}
\textbf{Sport} & \textbf{Community} & \textbf{Months} & \textbf{\# of} &  \textbf{Time Period} \\ 
 & & & \textbf{comments} & \\\hline

 Soccer& /r/soccer &51 & 3,449,070& Dec'14 - Feb'19 \\
 Football& /r/nfl& 51& 5,299,037&Dec'14 - Feb'19 \\
 Hockey& /r/nhl & 62& 32,815&Jan'15 - Feb'19 \\
Basketball & /r/nba & 51& 4,050,545& Dec'15 - Feb'19  \\\hline
\end{tabular}

\caption{Summary of the Sports Dataset}

\label{tab:sportsSummary}
\end{table}

\subsection{Thematically Similar Communities}
We now ask if our findings hold for a seemingly harder problem - communities stem from the same broad topic and same platform.
Data for four thematically similar  sports-based communities on reddit (/r/nhl, /r/nfl, /r/nba, /r/soccer) is described in Table \ref{tab:sportsSummary}. 

Again performance is excellent and not statistically different for style and content classifiers. 
Average accuracies and F-scores are 98.15\% and 0.982 for style and 98.72\% and 0.987 for content classifiers.
Thus our style classifier is capable of distinguishing between close communities.

\subsection{Platform Level Style}
Perhaps there are platform level (viewed as supra-communities) stylistic patterns that influence the member subreddits, subverses and 4chan boards.
The greater the stylistic influence of a platform on a community the higher the number of shared representative features likely.
We find that /4c/travel and /v/travel were stylistically most similar to their platforms since both shared 7 of their top-10 features with the parent community. In contrast, /r/television was stylistically the least similar to its parent reddit community with no shared feature.
The remaining 
six communities were also distinct, with 2 or less features shared with the parent platform-level community. 

Amongst the shared features we see that the use of `swear words' is representative of 4chan in general, and also of both /4c/politics and /4c/travel. While the use of agreeable words (`assent' category) was representative for the voat platform and both /v/politics and /v/travel. In contrast reddit did not have a platform level feature which was shared by more than a single sub-community.

\begin{table}[]
\scriptsize
\begin{tabular}{p{24.5mm}|p{12.7mm}|p{8.2mm}|p{8mm}}
\textbf{Category excluded}& \textbf{\# of features} & \textbf{F-Score} & \textbf{Accuracy} \\\hline
Parts of Speech &184& 96.02 & 96.68\% \\
Word level features w/o LIWC &241 & 95.02 & 95.84\% \\
Readability &256&  94.65 & 95.60\% \\
Character Level Features &219&  94.56 & 95.48\% \\
LIWC-$v3$ &148&  94.18 & 94.99\% \\
\textit{Content}&\textit{1,211,385}&\textit{92.20}&\textit{93.50\%}\\
\hline
\end{tabular}

\caption{Performance when feature category is excluded.}
\label{table:exclusive}
\end{table}

\begin{table}[]
\scriptsize
\begin{tabular}{p{24.5mm}|p{12.7mm}|p{8.2mm}|p{8mm}}
\textbf{Category included}& \textbf{\# of features} & \textbf{F-Score} & \textbf{Accuracy} \\\hline
All (LIWC-$v3$)& 262 & 95.20 & 96.08\% \\
All (LIWC-$v2$)& 282 &95.02 & 95.84\% \\\hline
Word level features (includes LIWC-$v3$)&  135  & 94.17 & 95.24 \\
LIWC-$v3$& 114 & 93.54 & 94.51\% \\
\textit{Content}&\textit{1,211,385}&\textit{92.20}&\textit{93.50\%}\\
Word level features w/o LIWC&21 & 90.80 & 92.10\% \\
Parts of Speech&78 & 90.49 & 91.78\% \\
Character Level Features&43 &  88.88 & 90.59\% \\
Readability&6 &  84.40 & 86.97\%\\\hline
\end{tabular}
\caption{Performance when the category of features listed is the only one included.}

\label{table:inclusive}
\end{table}

\subsection{Platform Level Readability Scores} 
Finally we explore nuances related language complexity at the platform level.  Ranking the three platforms by language complexity we observed that rankings were largely the same across the 7 readability measures we used (Table \ref{tab:featsAll}). From the most to least complex we have:
reddit, voat then 4chan.

The Dale-Chall readability measure differed slightly, ranking voat as having the most complex language followed by reddit then 4chan\footnote{The ARI and CLI rely on character count to measure language complexity. The Dale-Chall readability index, SMOG and the Gunning Fog index all measure the ratio of complex words to the total words. While, both SMOG and Gunning Fog define complexity based on the number of syllables,  Dale-Chall readability index defines complexity as based on a list of 3,000 predefined simple words. This difference in definitions and the small size of 3,000 explain why it gives a slightly different readability result.}.
We note that this ordering is fairly consistent with the complexity based ordering of the individual communities analyzed. Voat showing the most variation (relatively) indicates lower platform-level consistency.

\section{Limitations and Conclusions}

An individual's linguistic style is known to develop through subconscious processes while vocabulary to express content can be acquired through a deliberate conscious process.
We suggest that community-style is likely also acquired or shaped through a `subconscious' process that occurs through the interactions between community members.
In contrast to most of prior research, our communities of interest are made up of individuals interacting because of shared interests.
We find that communities have distinctive style and that we can use 200+ style features to successfully predict community membership. Additionally, style based classifiers are able to predict community membership as well as content-based classifiers.  

We made use of pseudo-documents as representative of the group's language, by doing so we assume that changes in style are minimal across time. 
Additionally, our study is limited to a few communities within 3 social media platforms. 
We will address these limitations in future research.

\bibliography{references}

\begin{thebibliography}{}

\bibitem[\protect\citeauthoryear{Abrams and Hogg}{2006}]{abrams2006social}
Abrams, D., and Hogg, M.~A.
\newblock 2006.
\newblock {\em Social identifications: A social psychology of intergroup
  relations and group processes}.
\newblock Routledge.

\bibitem[\protect\citeauthoryear{Ayuso}{2011}]{ayuso2011lucky}
Ayuso, M.~G.
\newblock 2011.
\newblock " how lucky for you that your tongue can taste the'r'in'parsley'":
  trauma theory and the literature of hispaniola.
\newblock {\em Afro-Hispanic Review}  47--62.

\bibitem[\protect\citeauthoryear{Bernstein \bgroup et al\mbox.\egroup
  }{2011}]{bernstein20114chan}
Bernstein, M.~S.; Monroy-Hern{\'a}ndez, A.; Harry, D.; Andr{\'e}, P.; Panovich,
  K.; and Vargas, G.
\newblock 2011.
\newblock 4chan and/b: An analysis of anonymity and ephemerality in a large
  online community.
\newblock In {\em Fifth International AAAI Conference on Weblogs and Social
  Media}.

\bibitem[\protect\citeauthoryear{Bucholtz and
  Hall}{2004}]{bucholtz2004language}
Bucholtz, M., and Hall, K.
\newblock 2004.
\newblock Language and identity.
\newblock {\em A companion to linguistic anthropology} 1:369--394.

\bibitem[\protect\citeauthoryear{Cheng, Chandramouli, and
  Subbalakshmi}{2011}]{cheng2011author}
Cheng, N.; Chandramouli, R.; and Subbalakshmi, K.
\newblock 2011.
\newblock Author gender identification from text.
\newblock {\em Digital Investigation} 8(1):78--88.

\bibitem[\protect\citeauthoryear{Coleman and Liau}{1975}]{coleman1975computer}
Coleman, M., and Liau, T.~L.
\newblock 1975.
\newblock A computer readability formula designed for machine scoring.
\newblock {\em Journal of Applied Psychology} 60(2):283.

\bibitem[\protect\citeauthoryear{Daelemans}{2013}]{daelemans2013explanation}
Daelemans, W.
\newblock 2013.
\newblock Explanation in computational stylometry.
\newblock In {\em International Conference on Intelligent Text Processing and
  Computational Linguistics},  451--462.
\newblock Springer.

\bibitem[\protect\citeauthoryear{Feng, Banerjee, and
  Choi}{2012}]{feng2012syntactic}
Feng, S.; Banerjee, R.; and Choi, Y.
\newblock 2012.
\newblock Syntactic stylometry for deception detection.
\newblock In {\em Proceedings of the 50th Annual Meeting of the Association for
  Computational Linguistics: Short Papers-Volume 2},  171--175.
\newblock Association for Computational Linguistics.

\bibitem[\protect\citeauthoryear{Gonzales, Hancock, and
  Pennebaker}{2010}]{gonzales2010language}
Gonzales, A.~L.; Hancock, J.~T.; and Pennebaker, J.~W.
\newblock 2010.
\newblock Language style matching as a predictor of social dynamics in small
  groups.
\newblock {\em Communication Research} 37(1):3--19.

\bibitem[\protect\citeauthoryear{Gunning}{1969}]{gunning1969fog}
Gunning, R.
\newblock 1969.
\newblock The fog index after twenty years.
\newblock {\em Journal of Business Communication} 6(2):3--13.

\bibitem[\protect\citeauthoryear{Hu \bgroup et al\mbox.\egroup
  }{2016}]{hu2016language}
Hu, T.; Xiao, H.; Luo, J.; and Nguyen, T.-v.~T.
\newblock 2016.
\newblock What the language you tweet says about your occupation.
\newblock In {\em Tenth International AAAI Conference on Web and Social Media}.

\bibitem[\protect\citeauthoryear{Hu, Talamadupula, and
  Kambhampati}{2013}]{hu2013dude}
Hu, Y.; Talamadupula, K.; and Kambhampati, S.
\newblock 2013.
\newblock Dude, srsly?: The surprisingly formal nature of twitter's language.
\newblock In {\em Seventh International AAAI Conference on Weblogs and Social
  Media}.

\bibitem[\protect\citeauthoryear{Pennebaker, Francis, and
  Booth}{2001}]{pennebaker2001linguistic}
Pennebaker, J.~W.; Francis, M.~E.; and Booth, R.~J.
\newblock 2001.
\newblock Linguistic inquiry and word count: Liwc 2001.
\newblock {\em Mahway: Lawrence Erlbaum Associates} 71(2001):2001.

\bibitem[\protect\citeauthoryear{Pennebaker}{2011}]{pennebaker2011secret}
Pennebaker, J.~W.
\newblock 2011.
\newblock The secret life of pronouns.
\newblock {\em New Scientist} 211(2828):42--45.

\bibitem[\protect\citeauthoryear{Potthast \bgroup et al\mbox.\egroup
  }{2017}]{potthast2017stylometric}
Potthast, M.; Kiesel, J.; Reinartz, K.; Bevendorff, J.; and Stein, B.
\newblock 2017.
\newblock A stylometric inquiry into hyperpartisan and fake news.
\newblock {\em arXiv preprint arXiv:1702.05638}.

\bibitem[\protect\citeauthoryear{Reynolds}{2018}]{reynolds_2018}
Reynolds, M.
\newblock 2018.
\newblock The wheels are falling off the alt-right's version of the internet.

\bibitem[\protect\citeauthoryear{Safin and Ogaltsov}{2018}]{safin2018detecting}
Safin, K., and Ogaltsov, A.
\newblock 2018.
\newblock Detecting a change of style using text statistics.
\newblock {\em Working Notes of CLEF}.

\bibitem[\protect\citeauthoryear{Sapkota \bgroup et al\mbox.\egroup
  }{2014}]{sapkota2014cross}
Sapkota, U.; Solorio, T.; Montes, M.; Bethard, S.; and Rosso, P.
\newblock 2014.
\newblock Cross-topic authorship attribution: Will out-of-topic data help?
\newblock In {\em Proceedings of COLING 2014, the 25th International Conference
  on Computational Linguistics: Technical Papers},  1228--1237.

\bibitem[\protect\citeauthoryear{Smucker, Allan, and
  Carterette}{2007}]{smucker2007comparison}
Smucker, M.~D.; Allan, J.; and Carterette, B.
\newblock 2007.
\newblock A comparison of statistical significance tests for information
  retrieval evaluation.
\newblock In {\em Proceedings of the sixteenth ACM conference on Conference on
  information and knowledge management},  623--632.
\newblock ACM.

\bibitem[\protect\citeauthoryear{Tausczik and
  Pennebaker}{2010}]{tausczik2010psychological}
Tausczik, Y.~R., and Pennebaker, J.~W.
\newblock 2010.
\newblock The psychological meaning of words: Liwc and computerized text
  analysis methods.
\newblock {\em Journal of language and social psychology} 29(1):24--54.

\bibitem[\protect\citeauthoryear{Van~Halteren \bgroup et al\mbox.\egroup
  }{2005}]{van2005new}
Van~Halteren, H.; Baayen, H.; Tweedie, F.; Haverkort, M.; and Neijt, A.
\newblock 2005.
\newblock New machine learning methods demonstrate the existence of a human
  stylome.
\newblock {\em Journal of Quantitative Linguistics} 12(1):65--77.

\bibitem[\protect\citeauthoryear{Wang \bgroup et al\mbox.\egroup
  }{2017}]{wang2017understanding}
Wang, Y.; Tang, J.; Li, J.; Li, B.; Wan, Y.; Mellina, C.; O'Hare, N.; and
  Chang, Y.
\newblock 2017.
\newblock Understanding and discovering deliberate self-harm content in social
  media.
\newblock In {\em Proceedings of the 26th International Conference on World
  Wide Web},  93--102.
\newblock International World Wide Web Conferences Steering Committee.

\bibitem[\protect\citeauthoryear{Zaman \bgroup et al\mbox.\egroup
  }{2019}]{zaman2019detecting}
Zaman, A.; Acharyya, R.; Kautz, H.; and Silenzio, V.
\newblock 2019.
\newblock Detecting low self-esteem in youths from web search data.
\newblock In {\em The World Wide Web Conference},  2270--2280.
\newblock ACM.

\end{thebibliography}
\bibliographystyle{aaai}
\label{EndOfPaper}
\end{document}